%% file: sample-ceur.tex
\documentclass[
twocolumn,
hf,
]{ceurart}

\usepackage{listings}
\usepackage{xspace}
\usepackage{booktabs}
\usepackage{paralist}
\usepackage{stfloats}
\usepackage{float}
\usepackage{enumitem}
\usepackage{etaremune}

\usepackage{soul}
\usepackage{xcolor}

\makeatletter
\newdimen\SOUL@dimen %new
\def\SOUL@ulunderline#1{{%
    \setbox\z@\hbox{#1}%
    \SOUL@dimen=\wd\z@ %new
    \dimen@i=\SOUL@uloverlap
    \advance\SOUL@dimen2\dimen@i %\dimen@ exchanged too
    \rlap{%
        \null
        \kern-\dimen@i
        \SOUL@ulcolor{\SOUL@ulleaders\hskip\SOUL@dimen}% new
    }%
    \unhcopy\z@
}}

\makeatother

\begin{document}

\copyrightyear{2022}
\copyrightclause{
  Copyright for this paper by its authors.
  Use permitted under Creative Commons License Attribution 4.0
  International (CC BY 4.0).}

\conference{
The AAAI-23 Workshop on Artificial Intelligence Safety (SafeAI 2023)
}

\title{Capabilities for Better ML Engineering}

% \tnotetext[1]{You can use this document as the template for preparing your
%   publication. We recommend using the latest version of the ceurart style.}

\author[1]{Chenyang Yang}
\address[1]{School of Computer Science, Carnegie Mellon University}

\author[2]{Rachel Brower-Sinning}
\address[2]{Carnegie Mellon Software Engineering Institute}

\author[2]{Grace A. Lewis}

\author[1]{Christian K\"astner}
% \cormark[1]

\author[1]{Tongshuang Wu}
% \cormark[1]

% \cortext[1]{Corresponding author.}

\begin{abstract}
% We envision a capability-based framework to unite existing efforts towards better ML engineering.
% We describe concrete scenarios to demonstrate capabilities’ broad applications across many different dimensions
% and how they help engineer safer ML models and systems in particular.
% Our preliminary results show capabilities' potential for improving ML engineering as well as highlight challenges of its adoption.
% We discuss challenges and opportunities for capabilities' integration into ML engineering.
In spite of machine learning’s rapid growth, its engineering support is scattered in many forms, and tends to favor certain engineering stages, stakeholders, and evaluation preferences.
We envision a capability-based framework, which uses fine-grained specifications for ML model behaviors to unite existing efforts towards better ML engineering.
We use concrete scenarios (model design, debugging, and maintenance) to articulate capabilities’ broad applications across various different dimensions, and their impact on building safer, more generalizable and more trustworthy models that reflect human needs.
Through preliminary experiments, we show the potential of capabilities for reflecting model generalizability, which can provide guidance for the ML engineering process.
We discuss challenges and opportunities for the integration of capabilities into ML engineering.
\end{abstract}


\maketitle

\input{body/macros}
\input{body/1-introduction}
\input{body/2-background}

\input{body/3-vision}
% \input{body/4-experiment}
\input{body/5-discussion}

\begin{acknowledgments}
% \paragraph{Acknowledgments.}
Kästner and Yang’s work is supported in part by 
NSF awards 1813598, 2131477, and 2206859 and support from the SEI. 
Lewis' and Brower-Sinning's work was funded and supported by the Department of Defense under Contract No. FA8702-15-D-0002 with Carnegie Mellon University for the operation of the Software Engineering Institute, a federally funded research and development center (DM22-1187).
% Wu's work
\end{acknowledgments}

%%
%% Define the bibliography file to be used
% \bibliographystyle{apa}
\bibliography{sample-ceur}

\appendix

\end{document}

%% file: body/macros.tex
\newcommand{\note}[2]{{\color{#1}{[#2]}}\xspace}
\newcommand{\sherry}[1]{\note{purple}{Sherry: #1}}
\newcommand{\cyang}[1]{\note{teal}{Chenyang: #1}}

\newcommand{\specialcell}[2][c]{%
  \begin{tabular}[#1]{@{}l@{}}#2\end{tabular}}

%% file: body/1-introduction.tex
\begin{table*}[H]
% \footnotesize
\caption{Example capabilities for pedestrian detection models.
Capabilities commonly express what a human would expect from ML models (common knowledge, robustness, human-style reasoning)
and can reflect different model qualities (generalizability, robustness, fairness).
We also illustrate possible instantiation strategies to produce concrete examples from capabilities.
}
\begin{tabular}{lll}
\toprule
{\textbf{Capability}} & {\textbf{Instantiation}} & \textbf{{Origin/Theory}} \\ \midrule
\specialcell[t]{Recognize pedestrians \\in wheelchairs} & \specialcell[t]{Curate images w/ pedestrians \\in wheelchairs} & Knowledge of important outliers \\
\addlinespace
% common target outlier
Robust to extreme weather & Transform sunny images to rainy & Robustness to anticipated distribution shift  \\
\addlinespace
% some distribution shift
Detect pedestrians of all ages & Slice test data by pedestrian age & Reasoning about concept variations \\
% some reasoning
\bottomrule
\end{tabular}~
% \vspace{-10pt}
\label{tab:capabilities}
\end{table*}

\section{Introduction}

% Despite many promising results when training and evaluating ML models, many ML projects fail when
% integrating ML models into production systems~\cite{consultants/news}.
% Many models are initially developed under idealized settings (e.g., with static datasets, following the i.i.d. assumption,
% assuming equal importance of all mistakes).

%The integration of machine learning (ML) models into production systems is challenging~\cite{icse22collaboration}
%% The process goes far beyond mere data and model engineering. 
%It requires engineering support in many different stages, including both model-level activities (e.g., data collection, model development, and model evaluation) and system-level activities (e.g., requirement engineering, system quality assurance, and model deployment).
%The whole process of building models and integrating them into a larger system is commonly referred to as ML engineering~\cite{burkov2020machine}.

Despite the rapid evolution of machine learning models, most effort has been on \emph{prototyping} models --- developing models under idealized settings (e.g., with static datasets, following the i.i.d. assumption, assuming equal importance of all mistakes).
These models tend to suffer in the wild where the ideal assumptions do not hold, leading to safety issues, fairness issues, and project failures~\cite{consultantsnews}. 
For example, a pedestrian detection model trained on images taken on sunny days would not correctly respond to natural weather changes~\cite{pedetrain09survey} and may have never seen a wheelchair user in training or test data.
Oversimplification has real consequences.
If we had only tested the aforementioned pedestrian detector on similar, sunny test examples, and used our overly optimistic evaluation to support deployment decisions, then an automated vehicle with the detector would be likely to cause accidents.

% In fact, we have witnessed safety concerns from a series of aspects --- e.g. models being unfair to certain populations~\cite{}, being vulnerable to attacks~\cite{}, etc.

To actually integrate models into production, substantial additional engineering effort is required by interdisciplinary teams~\cite{icse22collaboration}:
Not only do we need to make careful decisions at the model level (e.g., develop evaluation metrics that reflect human expectations on models~\cite{ribeiro-etal-2020-beyond}), but we also need to connect the model with the broader system design (e.g., the model functionalities should be well-specified in a \emph{requirements engineering process}~\cite{REbook}, similar to how we design user interfaces).

The importance of these efforts, commonly referred to as \textit{ML engineering}~\cite{burkov2020machine}, has been well-recognized, but the actual implementation tends to be scattered. 
For example, academic research on ML engineering tends to focus on the narrow space of model testing and debugging for data scientists~\cite[e.g.,][]{wu-etal-2019-errudite, ribeiro-lundberg-2022-adaptive}, whereas industrial efforts are mostly limited to supporting pipeline automation and model deployment (``MLOps'')~\cite{Mkinen2021MLOps}.
% aiming to help transition ML models to production systems.
More importantly, because these efforts are isolated, it is unclear how insights from one stage can be transferred to benefit the entire ML engineering process (e.g., how error analysis results help update model design decisions).
In other words, there is still a lack of synergy among existing efforts for better ML engineering practices.

In this work, we envision a unified framework for ML engineering.
In particular, we center our framework around \emph{capabilities}~\cite{ribeiro-etal-2020-beyond}.
A capability is a form of fine-grained specification for ML model behavior.
It helps define concrete model behaviors in various scenarios which are finer-grained and more holistic than standard evaluation metrics.
In our pedestrian detector example, different capabilities can be used to express safety requirements from different aspects, e.g., recognizing pedestrians in wheelchairs,
being robust to extreme weather, or being fair to people from different age groups~\cite{pedetrain09survey}.

Similar to other ML engineering efforts, the term capability emerged specifically from (and is mostly used in) model testing and debugging~\cite{ribeiro-etal-2020-beyond, ribeiro-lundberg-2022-adaptive}.
However, its natural link with \emph{expected model behaviors} makes it ideal for \emph{ML model specification} which, akin to software specification, (1) builds the root for the entire ML engineering cycle, going from model design all the way to deployment and maintenance, and (2) serves as the \emph{boundary object}~\cite{boundaryobject} for different stakeholders to negotiate their (sometimes conflicting) expectations of models.
Moreover, capabilities have the potential to reflect multiple essential factors in ML engineering, e.g., distribution shift~\cite{nips19shift}, robustness~\cite{goel-etal-2021-robustness}, fairness~\cite{shah-etal-2020-predictive} (see Tab.~\ref{tab:capabilities}).
However, capabilities have yet to fulfill their potential due to several challenges, e.g., it is not clear how to (1) best identify capabilities, (2) instantiate abstract capabilities, and (3) operationalize capabilities to maximize their utility.

We take the first step towards presenting the vision of a capability-based framework that both unites existing efforts and sheds light on future opportunities. 
Specifically, we illustrate the broad applicability of the framework from both the technical perspective and the practical perspective, by 
(1) summarizing how existing ML engineering concepts can be expressed with capabilities, and (2) describing four usage scenarios with unique characteristics (model debugging, collaboration, external quality assurance, and model maintenance).
We also conduct an exploratory study to demonstrate the feasibility of our vision. We conclude the paper by discussing challenges and opportunities for capabilities' integration into ML engineering that emerge from our preliminary results.

%% file: body/2-background.tex
\section{Capabilities}
\label{sec:background}

% \sherry{I think it's fair to connect much more closer to SE, and also articulate the limitations of current capabilities so you can motivate your work: 

% (1) ``capabilities'' emerges from the vision of ``testing AI models like testing softwares'' (this is true, in CheckList we said something like test ML models like testing SE, and that unit tests inspired MFT, Metamorphic test inspired INV, etc.); 

% (2) prior work has shown capability is nice for exposing some errors; 

% (3) but 
% 3a. its connection to SE is actually still shallow, for example requirement engineering is important for defining softwares but capabilities are not carefully selected right now in AI; And testing is shared between 

% 3b. AI has some diff from SE that makes its capability a bit unique, e.g. the definition of ``error'' is more vague (you can be 80\% correct), it cares much more about e.g. generalizability, etc. or whatever other things you have been discussing with Christian

% (4) This work tries to outline some actual use scenarios, and reflect better on the challenges and opportunities for maximizing the utility of capabilities.
% }

\paragraph{Capability definition: ML ``specification.''}
A capability can roughly be defined as a fine-grained specification of behaviors expected of an ML model.
The key idea is to go beyond just considering the overall accuracy of a model but analyzing to what degree
the model exhibits specific kinds of expected behaviors.
The term capability was popularized by work on testing specific behaviors of ML models~\cite{ribeiro-etal-2020-beyond},
but similar concepts can be found in other work on model testing (e.g., stress tests~\cite{naik-etal-2018-stress}) 
and in various work exploring nuances of model misbehavior and shortcut learning (e.g., underspecifications~\cite{underspecification}).
Previous work~\cite[e.g.,][]{ribeiro-etal-2020-beyond, ribeiro-lundberg-2022-adaptive} has shown that 
capabilities can expose many systematic problems in state-of-the-art models, are useful for interactive testing and debugging, and can guide data augmentation to train better models.
% We view capabilities as a powerful concept that could benefit the entire ML engineering process.

Capabilities share similarities with traditional \textit{software specifications} (and \textit{functional requirements}) in that both prescribe how software should behave in specific scenarios.
These prescriptions are general concepts or descriptions but can be concretized into a list of input-output examples (i.e., test cases) for assessing models in the engineering process.
% Capabilities specify how models should behave in different scenarios. 
We refer to the process of deriving test data from capabilities as \textit{instantiation}.
% To use abstract capabilities in concrete scenarios, we need to instantiate capabilities into concrete examples.
Capabilities can be instantiated in many different ways, including slicing existing data~\cite{wu-etal-2019-errudite}, transformation of existing data~\cite{nl-augmenter}, generating data from templates~\cite{ribeiro-etal-2020-beyond}, and targeted curation of new data (possibly with crowdsourcing)~\cite{Kaushik2020Learning} -- see examples in Tab.~\ref{tab:capabilities}.
% We could \textit{slice} an existing dataset to find the subpopulation corresponding to a capability.
% We could \textit{transform} existing data such that the predictions remain \textit{invariant}, or such that the predictions change in an expected way (often referred to as \textit{counterfactuals}~\cite{Kaushik2020Learning}).
% We could also \textit{curate} new data from templates, querying large models, or crowdsourcing.
Different instantiation strategies have different costs and benefits, and it is often necessary to make trade-offs between them.

However, capabilities also differ from traditional specifications in fundamental ways:
% \hl{
Traditional software is built using a \textit{deductive reasoning} process.
Their specifications are usually hard rules the software must satisfy -- a single input-output pair that violates the specification will be considered a bug.
In contrast, machine learning uses \textit{inductive reasoning}, where models are derived from observations and are expected to make occasional mistakes~\cite{mlisre}.
% \hl{
As such, instead of declaring a model as buggy for a single mistake related to a capability, we measure \emph{to what degree} the model has certain capabilities with a \emph{failure rate}.
In this sense, capability can be viewed as a soft \emph{lower bound specification}, and we use failure rates to look for issues where a model systematically underperforms with regard to a capability.
% }
% In traditional software testing, a single input-output pair that violates the specification is considered a bug, whereas ML models are expected to make occasional mistakes~\cite{mlisre}. 
% As such, instead of rejecting the model for a single mistake related to a capability, we measure \emph{to what degree} the model has certain capabilities with \emph{failure rate}.
% In this sense, capability can be viewed as a \emph{lower bound specification}, and we use failure rates to look for issues where a model systematically underperforms with regard to a capability.

% \sherry{Check this section -- I reordered stuff so similary goes together.}
%That is, we look for issues where a model systematically underperforms with regard to a capability.

% \hl{
% Another distinction is that traditional specifications are mostly used for \textit{deductive reasoning}.
% Software is built and tested to satisfy such specifications.
% In contrast, machine learning is an \textit{inductive reasoning} process, where hypotheses (i.e., models) are derived from observations.
% Therefore, capabilities 
% }

\paragraph{Capabilities as a unifying framework.}
There are many existing efforts to support ML engineering, but they are often scattered and unconnected.
Evaluating models on specific qualities like robustness, fairness, and generalizability is extensively discussed~\cite[e.g.,][]{ebrahimi-etal-2018-adversarial, shah-etal-2020-predictive, pmlr-v139-WILDS},
but they often focus exclusively on a narrow set of capabilities (e.g., robust to word replacement~\cite{sun22isotopic}, data shift~\cite{nips19shift}, and spurious correlations~\cite{mccoy-etal-2019-right}).
Different strategies for model evaluation and data augmentation, from slicing~\cite{wu-etal-2019-errudite}, counterfactuals~\cite{Kaushik2020Learning, gardner-etal-2020-contrast-set, wu-etal-2021-polyjuice}, templates~\cite{ribeiro-etal-2020-beyond}, to perturbations~\cite{nl-augmenter} are widely explored,
but there are very little efforts on combining them, evaluating their relative costs and effectiveness, and often such efforts are limited to individual qualities (e.g., robustness~\cite{goel-etal-2021-robustness}).
Recent work has shown interest in model debugging~\cite{ribeiro-lundberg-2022-adaptive, cscw21-crowdsourced-report} and error analysis~\cite{wu-etal-2019-errudite}, but they often use different terminologies despite the similar underlying ideas.

We argue that a \textit{capability} is a generic abstraction that can unify existing efforts. 
For example, 
different model evaluation strategies can be seen as ways to instantiate capabilities;
different model qualities can be viewed as (a series of) capabilities that might matter in specific scenarios;
a model's reliance on spurious correlations can be interpreted as a lack of specific capabilities (e.g., ignoring backgrounds for object detection~\cite{eccv18recog}).
Furthermore, as we will argue, capabilities can go beyond existing literature to benefit engineering stages (e.g., requirements engineering) and stakeholders (e.g., external evaluators or software engineers) that are currently under-explored.

%% file: body/3-vision.tex
\section{Capabilities for Better ML Engineering}
ML engineering effort happens at different development \textbf{stages}, with different \textbf{stakeholders} in the loop, and targets different model \textbf{qualities}. 
We argue that capabilities can help unify ML engineering efforts and lead to more systematic practice because they can play important roles in all these diverse dimensions.
% \begin{comment}
% \begin{itemize}[nosep, leftmargin=1.2em,labelwidth=*,align=left]
% \item Capabilities can be used at different \emph{stages} of ML engineering, including requirements engineering, data collection, model development, model evaluation, quality assurance, and model deployment. \sherry{minor, but this is weird that you just enumerated all the stages without going into any of them lol}
% \item Different \textbf{stakeholders} can benefit from capabilities. Data scientists use capabilities to debug and build better models~\cite{ribeiro-lundberg-2022-adaptive}; non-experts (e.g., external evaluators) use capabilities to reason about models and communicate with data scientists; end users build trust by understanding what models are capable of.
% \item Capabilities can relate to different \textbf{qualities} of ML models, ranging from accuracy, robustness, fairness, to generalizability.\sherry{similar to stages. Somehow stakeholder gets way more examples. Is this overview necessary at all? Maybe it's just about writing the first paragraph, and say we show how capability can be used in the four concrete scenarios that have different stage, stakeholder, and quality?}
% \end{itemize}
% \end{comment}

Below, we describe four concrete ML engineering scenarios (summarized in Tab.~\ref{tab:scenarios}), which cover different dimensions and highlight challenges and opportunities.

\begin{table}[t]
\footnotesize
\caption{Example usage scenarios for capabilities.
These scenarios cover different ML engineering stages and stakeholders,
showing capabilities are beneficial across dimensions.
}
\begin{tabular}{lll}
\toprule
{\textbf{Scenario}} & {\textbf{Stages}} & {\textbf{Stakeholders}} \\
\midrule 
Model Debugging &  Development & Data Scientists  \\
\addlinespace
Collaboration &  \specialcell[t]{Requirements, \\Evaluation} & \specialcell[t]{Software Engineers,\\Data Scientists}  \\
\addlinespace
External QA &  Evaluation & \specialcell[t]{External Evaluators,\\Regulators}          \\
\addlinespace
Model Maintenance &  Deployment   &\specialcell[t]{Data Scientists,\\End Users}   \\
\bottomrule
\end{tabular}
% \vspace{-10pt}
\label{tab:scenarios}
\end{table}

\subsection{Illustrative Scenarios}

\paragraph{Scenario 1: Model Debugging.}
Alice is a data scientist responsible for a chatbot used in her company. 
% \sherry{so is Alice debugging the chatbot? CY: YES.}
She is now debugging the conversational model that performs poorly on some inputs.
She tries to understand what is going wrong with these model mistakes. 
For each mistake, she speculates the potential issue behind it (e.g., input sentence contains numerical reasoning that the current model does not handle well) and updates the model accordingly.
% (e.g., through data augmentation). 
However, she finds the entire process ad-hoc and does not always produce a better model.
% However, after fixing these mistakes, she finds that the model performs slightly worse overall on the test set and new mistakes are introduced. 
% She has to go through the debugging process again and hopes to build a better model this time.

Capabilities can systematize this process and help Alice generalize from individual mistakes to systematic problems. 
Instead of chasing mistakes, Alice now \textit{identifies} common capabilities from model mistakes. 
Then she \textit{assesses} the importance of different capabilities, \textit{instantiates} the prioritized ones, and uses the instantiated tests for both training and evaluation. 
Alice now evaluates the new model not only on some general test data, but also on the test suites of different capabilities. 
She finds that the new model handles numerical reasoning better but is slightly worse on a different test suite that requires complex co-reference resolution. She decides that this is acceptable and releases the model.

\paragraph{Scenario 2: Collaboration.}
% [communications on what do not work]
% [cap could help describe target distribution without actually sharing the data]

Bob is a software engineer working in a government department, dealing with classified information. 
The department has a contract with an external data science team on a vision model for satellite images,
which is expected to be robust to various attacks and stable across various environments.
Due to strict data security policies, the external data science team relies on public datasets instead of actual production data. 
Bob struggles to communicate requirements and report useful feedback when the delivered model does not work in production.
% In particular, 
% Bob finds it hard to articulate their requirements to the data science team, without giving them access to private data.

% He finds out that the models perform much worse on their private datasets compared to what are reported by the data science team, 
% but he is not sure how to efficiently communicate with them without giving them access to private data.

Capabilities can serve as a \textit{communication} interface between different stakeholders. 
Bob would be able to clearly describe the failures in ways the data science team can understand, if he abstracts concrete private data, and identifies sharable \emph{capabilities} from them. 
Or even better, he can instantiate capabilities with public data points, such that the data science team can develop the next version of the model with a clear goal of improvement in mind in terms of capability failure rates.
% The data science team is expected to build models satisfying these capabilities, which are used for model development through data augmentation or customized model architecture.

% Alternatively, Bob could also come up with capabilities before model development. This is similar to traditional requirement engineering –  the data science team is expected to deliver a model that satisfies the requirements, i.e., achieve good performance on capability test suites.

\paragraph{Scenario 3: External Quality Assurance.}
Carolyn works for a quality assurance team that
previously focused on testing traditional software components. 
Carolyn is now responsible for independently evaluating models delivered by external contractors --- this time a model for fraud detection. 
Trained in traditional software testing, Carolyn finds it challenging to move forward without concrete specifications at hand, and is unsure what to do beyond standard accuracy evaluations.
%Carolyn finds it challenging to independently evaluate a model beyond just running standard accuracy evaluations on existing test data.
% She has access to the private dataset and needs to provide a thorough assessment of the model that are accessible to their customers.

Capabilities provide a more holistic view of how models perform in different scenarios. 
Carolyn reuses known capabilities for fraud detection, which her team developed for assessments on previous models, 
% \sherry{on this model? or on software? CY: model}
and evaluates the model on instantiated test suites from these capabilities, diving into specific capabilities of the model rather than providing just a single broad accuracy measure. 
She also looks at production data and past mistakes, and uses them to \textit{identify} new capabilities. 
Her final report \textit{communicates} how the model performs on different capability test suites and highlights the model’s major weaknesses.  

\paragraph{Scenario 4: Model Maintenance.}
Dan is a data scientist for a social media platform. They are responsible for a model that detects toxicity from user posts. 
The model performs well on previously curated data, but its performance degrades over time because of evolving trends in user posts. 
Dan tries to update the model periodically to cope with data shift. 
However, they find that the model is still frequently suboptimal to unknown future shifts even when trained with more recent data.

Capabilities can be used to track how data evolves through time and characterize data shift. 
Dan now maintains a list of high-quality capability test suites as regression tests. 
They regularly review new data to \textit{identify} whether the model needs additional capabilities, or whether the reliance on  existing capabilities changes over time.
This way, Dan gets to track the capability shift trajectory,  anticipate (to some extent) what future shift might look like, and can \textit{instantiate} suitable capabilities tests beforehand. 
With capabilities, Dan now builds and selects models that are more robust to data shift.

\paragraph{Discussion.}
% \sherry{Check -- I moved this part from the beginning to the end so (1) it's less repetitive (2) we can use the summary to reflect on the scenarios a bit more and end the section on a more impressive note.}
We described four different scenarios of using capabilities for better ML engineering, illustrating their broad applicability. As a recap,

\begin{itemize}[nosep, leftmargin=1.2em,labelwidth=*,align=left]
\item Capabilities can be used at different \textbf{stages} of ML engineering. On the one hand, they provide \emph{specifications} for ML models, which is fundamental to (collaborative) model design, development, and testing.
On the other hand, they also provide valuable abstractions for concrete data points, serve as a form of data specification, and allow for characterizing (possibly changing) deployment environments.
Notably, this potential for data documentation/specification further enlarges capabilities' impact on various stages that concern data, e.g., data collection, dataset evaluation, etc.

\item Different \textbf{stakeholders} can utilize capabilities. Though data scientists, external evaluators, etc. in our scenarios have different priorities in mind, they are able to converge on the capability framing --- whether to use capabilities to exploit their hypotheses on model mistakes, to communicate the characteristics of a non-shareable deployment environment, or to utilize prior training practices. Notably, as in the communication case, such convergence enables knowledge sharing or even negotiation between stakeholders, as everyone can speak the same ``language.''
\item Capabilities can relate to different \textbf{qualities} of ML models, ranging from accuracy (e.g., in \emph{debugging}), robustness (e.g., in \emph{collaboration}), fairness, to generalizability (e.g., in \emph{maintenance}). 
This enables multi-faceted evaluation without more consistent metric designs, which is valuable especially when multiple model qualities have to be balanced. 
\end{itemize}

Despite the promising future, these scenarios share common challenges, from identifying, assessing, communicating, to instantiating capabilities.
Yet different scenarios focus on different aspects and might have different requirements for the same challenge.
For example, all scenarios require identifying capabilities, but the ways they are identified or expressed vary; a \emph{shared language} would be required for collaboration, but if different stakeholders describe the same capabilities in different ways, or have different instantiation ideas, then additional inconsistency arises and has to be mitigated. 
We will discuss these practical barriers in the next section.

%\sherry{hold -- will get back after reading the next section}
% In model debugging, capabilities are usually identified from datasets, especially where the model makes mistakes, 
% while external evaluators might want to identify capabilities from domain knowledge ahead for quality assurance.
% For communication purposes, capabilities need to be expressed in a way that are intelligible to different stakeholders, 
% while in other scenarios like debugging and maintenance, they serve more as an internal documentation and could take forms in much more freedom.

\subsection{Exploratory Experiment}

To explore the practicality of our envisioned capability framework, we conducted an experiment to explore whether capabilities are reflective of model generalizability.
We focus on generalizability first because it is a primary design goal for any ML model, and a model quality essential for various use scenarios (e.g., the aforementioned model maintenance and collaboration). 
% but we plan to extend our experiments to cover other qualities and additional dimensions in the future.

%how useful capabilities are for model generalization, which is the focus in our model maintenance scenario, and is important for many other scenarios as well.

\paragraph{Experiment setup.}

\begin{table}[t]
\renewcommand{\arraystretch}{1.1}
\setlength{\tabcolsep}{3pt}

\footnotesize
\caption{Capabilities and their instantitation keywords for sentiment analysis, selected based on existing work~\cite{barnes-etal-2019-sentiment}. We slice the validation data on keywords to instantiate these capabilities, and the \% column represents the ratio of validation data that is included in the slice.
}
\begin{tabular}{r r p{0.67\linewidth}}
\toprule
\textbf{Capability}  & \% & \textbf{Keywords}\\
\midrule 
negation  & 51.6  &  not, n't  \\
negation \emph{(v2)} & 18.7 & no, never, neither, nobody, none, nor, nothing \\
shifter & 4.5 & refuse, reject, deny, doubt, abandon, miss, question, abort, stop \\
modality & 3.6 & would have, could have, should have \\
comparative & 16.6 & better, worse, than \\
mixed & 36.4 & but, however, though, although, despite, even if, rather than, except that \\
reducer & 14.1 & kind of, all that, less, a little, somewhat, still \\
amplifier & 48.8 & really, very, super, so, incredibly, extremely, at all, whatsoever, much \\
\bottomrule
\end{tabular}
\label{tab:cap-inst}
% \vspace{-10pt}
\end{table}

We define ``reflective'' as the statistical correlation between model performance on certain capability tests, and their performance on out-of-distribution data points.
% model generalizes under distribution shift.
\footnote{Experiment details can be found in an online appendix (\url{https://github.com/malusamayo/Capabilities-Experiment-Details}) and are not essential for the main vision outlined in this paper.}
% (details in Appendix~\ref{appendix:experiment} \sherry{I think I added enough details back so actually let's mode the appendix online, so all the spaces can be for the main paper lol}).

Specifically, in the experiment, we repeatedly finetuned BERT with different random seeds on the \textsc{Amazon-wilds} dataset~\cite{pmlr-v139-WILDS}, and obtained 100 sentiment analysis models with similar source domain accuracy (Amazon product reviews on \textsc{Home-and-kitchen}) but different target domain accuracy on 10 domains (e.g., \textsc{Movie-and-tv} reviews).
%We evaluated their accuracy on the source domain and multiple different target domains (e.g., Amazon product reviews, Tweets), where the accuracy on a target domain reflects how well the model generalizes under distribution shift.
We also selected eight capabilities for sentiment analysis from an existing study~\cite{barnes-etal-2019-sentiment}, and instantiated them into test suites through data slicing, as in Tab.~\ref{tab:cap-inst}.
%We collected a series of extra test suites, with guidance from capabilities (see Tab.~\ref{tab:cap-inst} for the list).
For each target domain, we fit a linear model to find correlations between the models' target domain accuracy (dependent variable), and the models' source domain accuracy, as well as their capability testing results (independent variables). 
We looked at adjusted $R^2$ to see whether the model has a better fit when these test results are part of the independent variable, compared to otherwise (i.e., whether these extra variables help predict out-of-distribution accuracy).
%We evaluated how well the models' accuracy on instantiated capability tests provide additional predictive power beyond accuracy in the source domain when predicting the models' accuracy on the target domain.

\paragraph{Results.}

\begin{figure}[t]
    \centering
    \includegraphics[width=\linewidth]{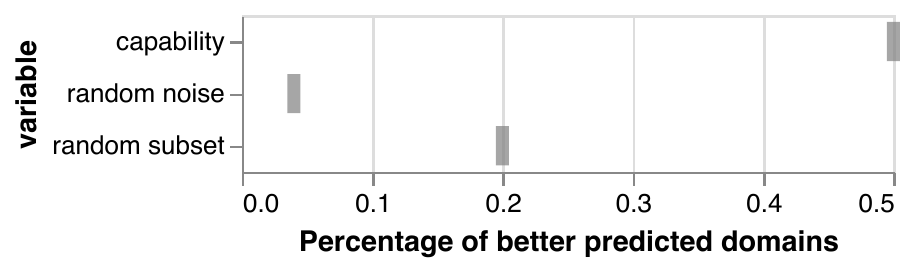}
    \vspace{-10pt}
    \caption{Capabilities better help predict model generalization compared to other baselines.}
    \label{fig:improvement}
    \vspace{10pt}

    \centering
    \includegraphics[trim={0cm 27cm 40cm 0cm}, clip,width=1\linewidth]{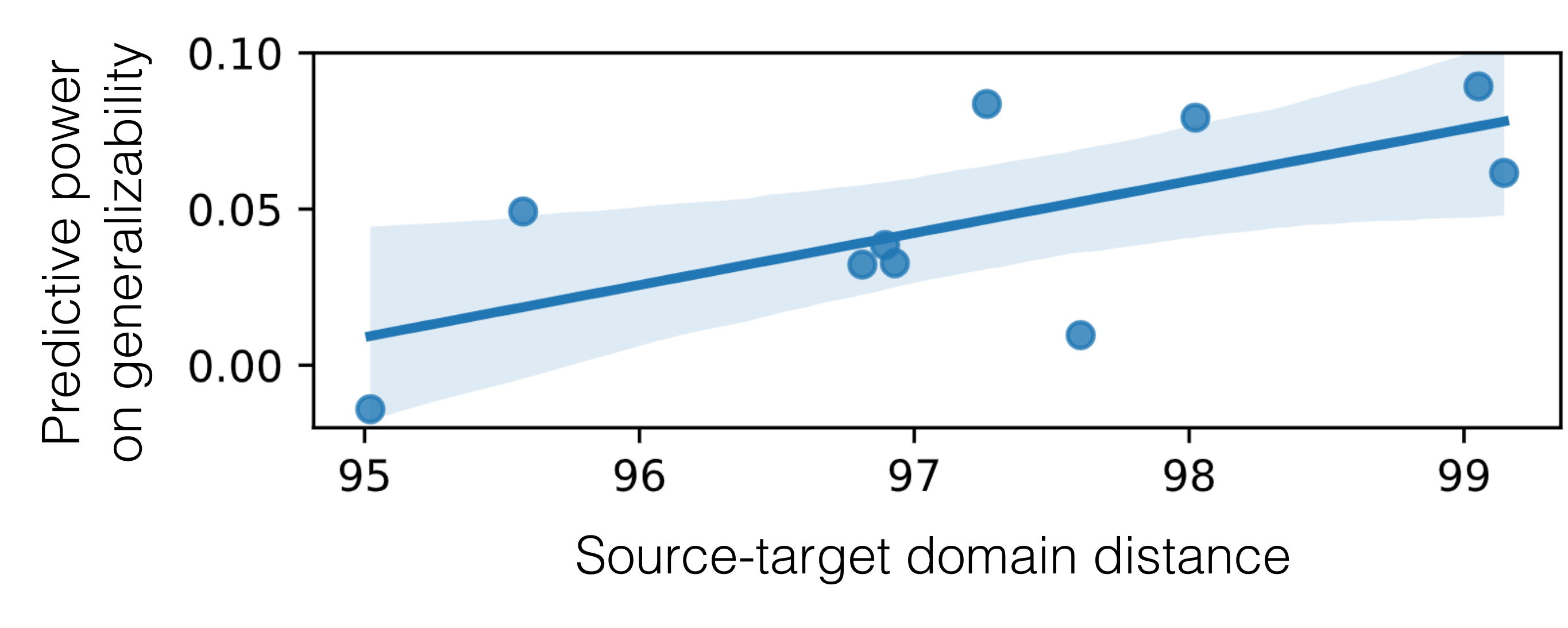}
    \vspace{-10pt}
    \caption{Predictive power improvement correlates with distribution distance.
    The further the distribution is, the better capabilities could help predict generalization. 
    We hypothesize that this is because if a target distribution is too close to the source distribution, there is little room left for improvement.
    }
    % \vspace{-10pt}
    \label{fig:distance}
\end{figure}

\emph{Model performance on capability tests is a strong signal for model’s generalizbability.}
We can confirm results from prior capability-testing experiments: Even fairly generic capabilities are somewhat helpful in predicting how well models generalize to out-of-distribution data.
In Fig.~\ref{fig:improvement}, on 50\% of the target domains (5/10), having capability tests adds a significant signal on models' generalizability to the target domain (i.e., significantly higher adjusted $R^2$).
In contrast, baselines with model performances on randomly sliced subsets or random noise do not provide similar improvement.

% This provides further evidence that capabilities are useful beyond the explored use cases in testing and debugging.
\emph{Capability tests especially helps predict how well models generalize to further domains.}
We also mapped out the approximated distances between each target domain and the source domain, using a proxy $\mathcal{A}$-distance~\cite{blitzer-etal-2007-biographies}.
As in Fig.~\ref{fig:distance}, we observe a positive slope between the distance and the $R^2$ power.
%capability-based tests are more indicative the larger the distance between source and target distributions (see  
This shows that capabilities are particularly helpful for distributions that deviate more from the training distribution, such as in Bob's scenario where distribution details could not be shared.

\paragraph{Discussion.}
Besides the positive signals, our experiment also highlighted several challenges we faced when using capabilities during ML engineering. 
In particular, we observe that vanilla capability identification and instantiation have limited utility, for two reasons:

\begin{itemize}[nosep, leftmargin=1.2em,labelwidth=*,align=left]
\item \textit{Different capabilities add different amount of information.}
Some capabilities (e.g., \textit{negation}) produce too many test cases, 
% \sherry{Maybe add a column in table 3 on number of examples per slice?}
which leads to an uninformative distribution close to the source dataset, while others (e.g., \textit{modality}) result in a rather distinct distribution, which is more informative for predicting generalization. 

\item \textit{Different capabilities add different kinds of information.}
Some capabilities are complementary but others are highly correlated and add little additional information over other capabilities.
For example, in our experiment, we found that using only \textit{shifter} improves predicting generalizability in 20\% cases, but adding \textit{modality} further improves in 40\% cases.
At the same time, capabilities could also add conflicting information, where models perform or generalize worse if they better support a capability, which is similar to common tradeoffs between model accuracy and other qualities (e.g., robustness).
% \sherry{Oh I remember some task-specific / agnostic tests like ``hot'' etc. If we used them in the experiment, we can discuss it here, or maybe somewhere in the next section? Or if it's a stretch then we don't have to.}
\end{itemize}

In essence, the design space of capabilities and their corresponding instantiations is massive.
While prior work has reported positive impacts of capabilities, as well as success in scaffolding the identification and instantiation process~\cite{ribeiro-etal-2020-beyond}, few studies have comprehensively evaluated the information gain of different capabilities, the interactions between capabilities, and the effectiveness of different identification / instantiation strategies. 

In our experiment, we resolved to the most basic and typical methods for identification and instantiation, which has inherent limitations:
We identified capabilities by reusing domain knowledge from existing work, which is not tailored for generalization to specific target distribution; 
we instantiated capabilities through coarse-grained slicing on keywords, which does not always produce useful test suites (e.g., \textit{negation}).
While we also considered other identification and instantiation strategies, we eventually discarded them as they require much more manual effort --- a reflection on the reality that most people would probably prefer simpler (if rather flawed) methods.

As a result, we argue that proper guidance needs to be designed, such that different stakeholders can quickly climb the rather steep learning curve for making capabilities useful. We discuss future directions next.

% \sherry{
% so do you want this to be a separate section or a sub-section?

% And I think we can write it to be:

% 1. We use some preliminary results to show the feasibility of using capabilities for essential ML engineering steps

% 2. We focus on generalization because it's related to multiple scenarios (see comments in table 1 caption)

% 3. We show capability tests:

% 3a. is predictive to model generalizability -> would help in debugging, doing better on capabilites actually mean (v.s. checklist only shows capabilities can expose model problems but didnt build the relation of better capablity -> better generalization)

% 3b. their predictiveness correlates with dataset distance -> would be useful for quality assurance for downstream deployment in particular domains

% 3c. Multiple capabilities add more information -> communication \& negotiation among multiple stakeholders

% 4. However, there are:

% 4a. uninformative capabilities that would just be removed because they overlap too much with dev set 

% 4b. Task specific vs. task agnostic capabilities

% -> capability identification is important and instantiation is also important

% -> Next section talks more about challenges and opportunities
% }

%% file: body/5-discussion.tex
\section{Challenges and Opportunities}
\label{sec:challenges}

To more systematically use capabilities, further research is needed. 
We argue that ML engineering can generally benefit from software engineering disciplines, with principles from requirements engineering and software testing in particular.
In the following, we identify promising research directions based on gaps in the literature and our own observations in our experiment.

% \hl{Connect better to SE literature}

% \looseness=-1
\paragraph{Identifying capabilities.}
It is challenging to identify capabilities for concrete scenarios.
Capabilities often differ across different modes (vision vs. language), different tasks (sentiment analysis vs. natural language inference), and different domains (product reviews vs. book reviews). 
While we may develop a catalog of common capabilities for general-purpose tasks, such as sentiment analysis~\cite{barnes-etal-2019-sentiment}, 
we will likely need to identify specific capabilities for each domain-specific problem.
% (e.g., capabilities of detecting toxicity towards immigrants for toxicity detection models).
Existing strategies include using domain knowledge~\cite{nl-augmenter}, performing error analysis~\cite{naik-etal-2018-stress, wu-etal-2019-errudite, cscw21-crowdsourced-report}, and mining knowledge from existing corpora~\cite{improvingdomainspecs}.
Most strategies require extensive efforts of domain experts or crowdsource workers, making them hard to scale.
% They are also often conducted in an unsystematic way, while they could have learned from discussion on how to identify software specifications in requirement engineering.
% \hl{
They are also often conducted in an unsystematic way and do not draw on classic requirement elicitation and participatory design approaches.
% }
Future work could explore:
% Existing knowledge base or large language models could be exploited for capability identification in an automated system or with human in the loop.
% Better interaction mechanisms and tool supports are another promising direction, especially when capabilities are identified from error analysis -- users need to efficiently interact with models for abstracting concrete errors into capabilities.
\begin{compactitem}
    \item[\textbf{RQ1}] How could we support more effective discovery and reuse of domain knowledge? When and how can we automate discovery?
    \item[\textbf{RQ2}] What kinds of mechanisms could support more efficient human-AI interaction in error analysis?
    \item[\textbf{RQ3}] How could we design a better process to help both experts and non-experts identify capabilities? 
    % \sherry{maybe just non-expert stakeholders? I thought domain knowledge / domain expert is in RQ1?}
\end{compactitem}

\paragraph{Assessing capabilities.}
% Different models may learn different capabilities and different capabilities are often encoded in existing data to different extents. 
% Future works could explore different metric and method designs that help evaluate and rank identified capabilities.

Capabilities often exhibit a hierarchical structure. 
For example, \textit{understanding negation} is a very general capability, whereas \textit{understanding double negation} or \textit{handling modifiers} as ``hardly'' and ``never'' are more specific (sub-)capabilities.
% or styles (e.g., variants of negative words on the Internet).
How fine-grained a capability should be will likely depend on the specific scenarios.
More coarse capabilities are more reusable, whereas finer-grained ones capture concrete concepts that might be especially useful for the domain (but may not transfer --- e.g., concrete adjectives like ``cold'' is positive when describing refrigerators but not so much for thermos).
Their predictiveness also differs across scenarios, as we observed in our experiments.
When identifying capabilities, we need to determine the proper granularity, and evaluate their importance within the context:

% Therefore, evaluation metrics and methods for capabilities are expected to be contextualized and augmented with human supervision.

\begin{compactitem}
    \item[\textbf{RQ4}] What is a good granularity for a capability?
    \item[\textbf{RQ5}] How do we evaluate/rank capabilities by context?
\end{compactitem}

% movie reviews, reddit comments, or medical records might re

\paragraph{Communicating capabilities.}
Identified capabilities need to be efficiently communicated between different stakeholders,
who might have different requirements and potential conflicts, or may describe the same capabilities in drastically different ways depending on their expertise (e.g., an expert may say ``invariant to environmental conditions'' when a lay user says ``performs the same in sunny, raining, stormy weathers.'')
Common communication vocabularies and conflict resolution mechanisms, possibly informed by existing requirements engineering literature, would greatly facilitate the process.
% where languages for requirements have been well explored and discussed.

\begin{compactitem}
    \item[\textbf{RQ6}] How can we develop a shared language or interface to facilitate capability communication?
    \item[\textbf{RQ7}] How can capabilities support conflict resolution between different stakeholders?
\end{compactitem}

\paragraph{Instantiating capabilities.}
Abstract capabilities need to be instantiated as concrete test cases,
to be further used as regression tests, examples for communication, or augmentation data for training.
Existing work has explored different strategies for instantiating capabilities (c.f.~Sec.~\ref{sec:background}),
but it remains unclear how different strategies perform in different scenarios and whether they could be combined in a meaningful way.
These strategies are similar to software testing (e.g., unit tests and metamorphic testing~\cite{MTsurvey18}) and can be informed by existing software engineering literature (e.g., test case generation, fuzzing, prioritization, and requirements validation).
\begin{compactitem}
    \item[\textbf{RQ8}] How should we select instantiation strategies in different scenarios? How to measure and trade off costs and benefits?
    \item[\textbf{RQ9}] How do different instantiation strategies complement each other?
\end{compactitem}

\section{Conclusion}
% \hl{
A capability is a generic abstraction that unifies existing efforts on model testing, debugging, and evaluation.
It can also benefit the entire ML engineering lifecycle from data collection to model deployment,
addressing the needs of different stakeholders and model qualities.
Our exploratory experiments showed that capabilities could provide strong signals for model generalizability, as well as highlighted challenges in integrating them into the ML engineering process.
We hope future research will better support identifying, assessing, communicating, and instantiating capabilities.
% }zz

%% file: sample-ceur.bbl
\begin{thebibliography}{30}
\expandafter\ifx\csname natexlab\endcsname\relax\def\natexlab#1{#1}\fi
\providecommand{\url}[1]{\texttt{#1}}
\providecommand{\href}[2]{#2}
\providecommand{\path}[1]{#1}
\providecommand{\DOIprefix}{doi:}
\providecommand{\ArXivprefix}{arXiv:}
\providecommand{\URLprefix}{URL: }
\providecommand{\Pubmedprefix}{pmid:}
\providecommand{\doi}[1]{\href{http://dx.doi.org/#1}{\path{#1}}}
\providecommand{\Pubmed}[1]{\href{pmid:#1}{\path{#1}}}
\providecommand{\bibinfo}[2]{#2}
\ifx\xfnm\relax \def\xfnm[#1]{\unskip,\space#1}\fi
%Type = Article
\bibitem[{Panetta(2020)}]{consultantsnews}
\bibinfo{author}{K.~Panetta},
\newblock \bibinfo{title}{Gartner identifies the top strategic technology
  trends for 2021.}  (\bibinfo{year}{2020}).
%Type = Article
\bibitem[{Gerónimo et~al.(2010)Gerónimo, López, Sappa, and
  Graf}]{pedetrain09survey}
\bibinfo{author}{D.~Gerónimo}, \bibinfo{author}{A.~M. López},
  \bibinfo{author}{A.~D. Sappa}, \bibinfo{author}{T.~Graf},
\newblock \bibinfo{title}{Survey of pedestrian detection for advanced driver
  assistance systems},
\newblock \bibinfo{journal}{IEEE Transactions on Pattern Analysis and Machine
  Intelligence} \bibinfo{volume}{32} (\bibinfo{year}{2010})
  \bibinfo{pages}{1239--1258}.
%Type = Inproceedings
\bibitem[{Nahar et~al.(2022)Nahar, Zhou, Lewis, and
  Kästner}]{icse22collaboration}
\bibinfo{author}{N.~Nahar}, \bibinfo{author}{S.~Zhou},
  \bibinfo{author}{G.~Lewis}, \bibinfo{author}{C.~Kästner},
\newblock \bibinfo{title}{Collaboration challenges in building ml-enabled
  systems: Communication, documentation, engineering, and process},
\newblock in: \bibinfo{booktitle}{2022 IEEE/ACM 44th International Conference
  on Software Engineering (ICSE)}, \bibinfo{year}{2022}, pp.
  \bibinfo{pages}{413--425}.
%Type = Inproceedings
\bibitem[{Ribeiro et~al.(2020)Ribeiro, Wu, Guestrin, and
  Singh}]{ribeiro-etal-2020-beyond}
\bibinfo{author}{M.~T. Ribeiro}, \bibinfo{author}{T.~Wu},
  \bibinfo{author}{C.~Guestrin}, \bibinfo{author}{S.~Singh},
\newblock \bibinfo{title}{Beyond accuracy: Behavioral testing of {NLP} models
  with {C}heck{L}ist},
\newblock in: \bibinfo{booktitle}{Proceedings of the 58th Annual Meeting of the
  Association for Computational Linguistics}, \bibinfo{publisher}{Association
  for Computational Linguistics}, \bibinfo{address}{Online},
  \bibinfo{year}{2020}, pp. \bibinfo{pages}{4902--4912}.
%Type = Book
\bibitem[{van Lamsweerde(2009)}]{REbook}
\bibinfo{author}{A.~van Lamsweerde}, \bibinfo{title}{Requirements Engineering:
  From System Goals to UML Models to Software Specifications},
  \bibinfo{edition}{1st} ed., \bibinfo{publisher}{Wiley Publishing},
  \bibinfo{year}{2009}.
%Type = Book
\bibitem[{Burkov(2020)}]{burkov2020machine}
\bibinfo{author}{A.~Burkov}, \bibinfo{title}{Machine learning engineering},
  volume~\bibinfo{volume}{1}, \bibinfo{publisher}{True Positive Incorporated},
  \bibinfo{year}{2020}.
%Type = Inproceedings
\bibitem[{Wu et~al.(2019)Wu, Ribeiro, Heer, and Weld}]{wu-etal-2019-errudite}
\bibinfo{author}{T.~Wu}, \bibinfo{author}{M.~T. Ribeiro},
  \bibinfo{author}{J.~Heer}, \bibinfo{author}{D.~Weld},
\newblock \bibinfo{title}{{E}rrudite: Scalable, reproducible, and testable
  error analysis},
\newblock in: \bibinfo{booktitle}{Proceedings of the 57th Annual Meeting of the
  Association for Computational Linguistics}, \bibinfo{publisher}{Association
  for Computational Linguistics}, \bibinfo{address}{Florence, Italy},
  \bibinfo{year}{2019}, pp. \bibinfo{pages}{747--763}.
%Type = Inproceedings
\bibitem[{Ribeiro and Lundberg(2022)}]{ribeiro-lundberg-2022-adaptive}
\bibinfo{author}{M.~T. Ribeiro}, \bibinfo{author}{S.~Lundberg},
\newblock \bibinfo{title}{Adaptive testing and debugging of {NLP} models},
\newblock in: \bibinfo{booktitle}{Proceedings of the 60th Annual Meeting of the
  Association for Computational Linguistics (Volume 1: Long Papers)},
  \bibinfo{publisher}{Association for Computational Linguistics},
  \bibinfo{address}{Dublin, Ireland}, \bibinfo{year}{2022}, pp.
  \bibinfo{pages}{3253--3267}.
%Type = Article
\bibitem[{M{\"a}kinen et~al.(2021)M{\"a}kinen, Skogstr{\"o}m, Laaksonen, and
  Mikkonen}]{Mkinen2021MLOps}
\bibinfo{author}{S.~M{\"a}kinen}, \bibinfo{author}{H.~Skogstr{\"o}m},
  \bibinfo{author}{E.~Laaksonen}, \bibinfo{author}{T.~Mikkonen},
\newblock \bibinfo{title}{Who needs mlops: What data scientists seek to
  accomplish and how can mlops help?},
\newblock \bibinfo{journal}{2021 IEEE/ACM 1st Workshop on AI Engineering -
  Software Engineering for AI (WAIN)}  (\bibinfo{year}{2021})
  \bibinfo{pages}{109--112}.
%Type = Inbook
\bibitem[{Star(1989)}]{boundaryobject}
\bibinfo{author}{S.~L. Star}, \bibinfo{title}{The Structure of Ill-Structured
  Solutions: Boundary Objects and Heterogeneous Distributed Problem Solving},
  \bibinfo{publisher}{Morgan Kaufmann Publishers Inc.}, \bibinfo{address}{San
  Francisco, CA, USA}, \bibinfo{year}{1989}, p. \bibinfo{pages}{37–54}.
%Type = Inbook
\bibitem[{Rabanser et~al.(2019)Rabanser, G\"{u}nnemann, and
  Lipton}]{nips19shift}
\bibinfo{author}{S.~Rabanser}, \bibinfo{author}{S.~G\"{u}nnemann},
  \bibinfo{author}{Z.~C. Lipton}, \bibinfo{title}{Failing Loudly: An Empirical
  Study of Methods for Detecting Dataset Shift}, \bibinfo{publisher}{Curran
  Associates Inc.}, \bibinfo{address}{Red Hook, NY, USA}, \bibinfo{year}{2019}.
%Type = Inproceedings
\bibitem[{Goel et~al.(2021)Goel, Rajani, Vig, Taschdjian, Bansal, and
  R{\'e}}]{goel-etal-2021-robustness}
\bibinfo{author}{K.~Goel}, \bibinfo{author}{N.~F. Rajani},
  \bibinfo{author}{J.~Vig}, \bibinfo{author}{Z.~Taschdjian},
  \bibinfo{author}{M.~Bansal}, \bibinfo{author}{C.~R{\'e}},
\newblock \bibinfo{title}{Robustness gym: Unifying the {NLP} evaluation
  landscape},
\newblock in: \bibinfo{booktitle}{Proceedings of the 2021 Conference of the
  North American Chapter of the Association for Computational Linguistics:
  Human Language Technologies: Demonstrations}, \bibinfo{publisher}{Association
  for Computational Linguistics}, \bibinfo{address}{Online},
  \bibinfo{year}{2021}, pp. \bibinfo{pages}{42--55}.
%Type = Inproceedings
\bibitem[{Shah et~al.(2020)Shah, Schwartz, and
  Hovy}]{shah-etal-2020-predictive}
\bibinfo{author}{D.~S. Shah}, \bibinfo{author}{H.~A. Schwartz},
  \bibinfo{author}{D.~Hovy},
\newblock \bibinfo{title}{Predictive biases in natural language processing
  models: A conceptual framework and overview},
\newblock in: \bibinfo{booktitle}{Proceedings of the 58th Annual Meeting of the
  Association for Computational Linguistics}, \bibinfo{publisher}{Association
  for Computational Linguistics}, \bibinfo{address}{Online},
  \bibinfo{year}{2020}, pp. \bibinfo{pages}{5248--5264}.
%Type = Inproceedings
\bibitem[{Naik et~al.(2018)Naik, Ravichander, Sadeh, Rose, and
  Neubig}]{naik-etal-2018-stress}
\bibinfo{author}{A.~Naik}, \bibinfo{author}{A.~Ravichander},
  \bibinfo{author}{N.~Sadeh}, \bibinfo{author}{C.~Rose},
  \bibinfo{author}{G.~Neubig},
\newblock \bibinfo{title}{Stress test evaluation for natural language
  inference},
\newblock in: \bibinfo{booktitle}{Proceedings of the 27th International
  Conference on Computational Linguistics}, \bibinfo{publisher}{Association for
  Computational Linguistics}, \bibinfo{address}{Santa Fe, New Mexico, USA},
  \bibinfo{year}{2018}, pp. \bibinfo{pages}{2340--2353}.
%Type = Misc
\bibitem[{D'Amour et~al.(2020)}]{underspecification}
\bibinfo{author}{A.~D'Amour}, et~al., \bibinfo{title}{Underspecification
  presents challenges for credibility in modern machine learning},
  \bibinfo{year}{2020}.
%Type = Misc
\bibitem[{Dhole et~al.(2021)}]{nl-augmenter}
\bibinfo{author}{K.~D. Dhole}, et~al., \bibinfo{title}{Nl-augmenter: A
  framework for task-sensitive natural language augmentation},
  \bibinfo{year}{2021}.
%Type = Inproceedings
\bibitem[{Kaushik et~al.(2020)Kaushik, Hovy, and Lipton}]{Kaushik2020Learning}
\bibinfo{author}{D.~Kaushik}, \bibinfo{author}{E.~Hovy},
  \bibinfo{author}{Z.~Lipton},
\newblock \bibinfo{title}{Learning the difference that makes a difference with
  counterfactually-augmented data},
\newblock in: \bibinfo{booktitle}{International Conference on Learning
  Representations}, \bibinfo{year}{2020}.
%Type = Misc
\bibitem[{Kaestner(2020)}]{mlisre}
\bibinfo{author}{C.~Kaestner}, \bibinfo{title}{Machine learning is requirements
  engineering — on the role of bugs, verification, and validation in machine
  learning}, \bibinfo{howpublished}{Blog}, \bibinfo{year}{2020}.
%Type = Inproceedings
\bibitem[{Ebrahimi et~al.(2018)Ebrahimi, Lowd, and
  Dou}]{ebrahimi-etal-2018-adversarial}
\bibinfo{author}{J.~Ebrahimi}, \bibinfo{author}{D.~Lowd},
  \bibinfo{author}{D.~Dou},
\newblock \bibinfo{title}{On adversarial examples for character-level neural
  machine translation},
\newblock in: \bibinfo{booktitle}{Proceedings of the 27th International
  Conference on Computational Linguistics}, \bibinfo{publisher}{Association for
  Computational Linguistics}, \bibinfo{address}{Santa Fe, New Mexico, USA},
  \bibinfo{year}{2018}, pp. \bibinfo{pages}{653--663}.
%Type = Inproceedings
\bibitem[{Koh et~al.(2021)}]{pmlr-v139-WILDS}
\bibinfo{author}{P.~W. Koh}, et~al.,
\newblock \bibinfo{title}{Wilds: A benchmark of in-the-wild distribution
  shifts},
\newblock in: \bibinfo{editor}{M.~Meila}, \bibinfo{editor}{T.~Zhang} (Eds.),
  \bibinfo{booktitle}{Proceedings of the 38th International Conference on
  Machine Learning}, volume \bibinfo{volume}{139} of
  \textit{\bibinfo{series}{Proceedings of Machine Learning Research}},
  \bibinfo{publisher}{PMLR}, \bibinfo{year}{2021}, pp.
  \bibinfo{pages}{5637--5664}.
%Type = Inproceedings
\bibitem[{Sun et~al.(2022)Sun, Zhang, Xiong, Harman, Papadakis, and
  Zhang}]{sun22isotopic}
\bibinfo{author}{Z.~Sun}, \bibinfo{author}{J.~M. Zhang},
  \bibinfo{author}{Y.~Xiong}, \bibinfo{author}{M.~Harman},
  \bibinfo{author}{M.~Papadakis}, \bibinfo{author}{L.~Zhang},
\newblock \bibinfo{title}{Improving machine translation systems via isotopic
  replacement},
\newblock in: \bibinfo{booktitle}{Proceedings of the 44th International
  Conference on Software Engineering}, ICSE '22,
  \bibinfo{publisher}{Association for Computing Machinery},
  \bibinfo{address}{New York, NY, USA}, \bibinfo{year}{2022}, p.
  \bibinfo{pages}{1181–1192}.
%Type = Inproceedings
\bibitem[{McCoy et~al.(2019)McCoy, Pavlick, and Linzen}]{mccoy-etal-2019-right}
\bibinfo{author}{T.~McCoy}, \bibinfo{author}{E.~Pavlick},
  \bibinfo{author}{T.~Linzen},
\newblock \bibinfo{title}{Right for the wrong reasons: Diagnosing syntactic
  heuristics in natural language inference},
\newblock in: \bibinfo{booktitle}{Proceedings of the 57th Annual Meeting of the
  Association for Computational Linguistics}, \bibinfo{publisher}{Association
  for Computational Linguistics}, \bibinfo{address}{Florence, Italy},
  \bibinfo{year}{2019}, pp. \bibinfo{pages}{3428--3448}.
%Type = Inproceedings
\bibitem[{Gardner et~al.(2020)}]{gardner-etal-2020-contrast-set}
\bibinfo{author}{M.~Gardner}, et~al.,
\newblock \bibinfo{title}{Evaluating models{'} local decision boundaries via
  contrast sets},
\newblock in: \bibinfo{booktitle}{Findings of the Association for Computational
  Linguistics: EMNLP 2020}, \bibinfo{publisher}{Association for Computational
  Linguistics}, \bibinfo{address}{Online}, \bibinfo{year}{2020}, pp.
  \bibinfo{pages}{1307--1323}.
%Type = Inproceedings
\bibitem[{Wu et~al.(2021)Wu, Ribeiro, Heer, and Weld}]{wu-etal-2021-polyjuice}
\bibinfo{author}{T.~Wu}, \bibinfo{author}{M.~T. Ribeiro},
  \bibinfo{author}{J.~Heer}, \bibinfo{author}{D.~Weld},
\newblock \bibinfo{title}{Polyjuice: Generating counterfactuals for explaining,
  evaluating, and improving models},
\newblock in: \bibinfo{booktitle}{Proceedings of the 59th Annual Meeting of the
  Association for Computational Linguistics and the 11th International Joint
  Conference on Natural Language Processing (Volume 1: Long Papers)},
  \bibinfo{publisher}{Association for Computational Linguistics},
  \bibinfo{address}{Online}, \bibinfo{year}{2021}, pp.
  \bibinfo{pages}{6707--6723}.
%Type = Article
\bibitem[{Cabrera et~al.(2021)Cabrera, Druck, Hong, and
  Perer}]{cscw21-crowdsourced-report}
\bibinfo{author}{A.~A. Cabrera}, \bibinfo{author}{A.~J. Druck},
  \bibinfo{author}{J.~I. Hong}, \bibinfo{author}{A.~Perer},
\newblock \bibinfo{title}{Discovering and validating ai errors with
  crowdsourced failure reports},
\newblock \bibinfo{journal}{Proc. ACM Hum.-Comput. Interact.}
  \bibinfo{volume}{5} (\bibinfo{year}{2021}).
%Type = Inproceedings
\bibitem[{Beery et~al.(2018)Beery, Van~Horn, and Perona}]{eccv18recog}
\bibinfo{author}{S.~Beery}, \bibinfo{author}{G.~Van~Horn},
  \bibinfo{author}{P.~Perona},
\newblock \bibinfo{title}{Recognition in terra incognita},
\newblock in: \bibinfo{booktitle}{Computer Vision – ECCV 2018: 15th European
  Conference, Munich, Germany, September 8-14, 2018, Proceedings, Part XVI},
  \bibinfo{publisher}{Springer-Verlag}, \bibinfo{address}{Berlin, Heidelberg},
  \bibinfo{year}{2018}, p. \bibinfo{pages}{472–489}.
%Type = Inproceedings
\bibitem[{Barnes et~al.(2019)Barnes, {\O}vrelid, and
  Velldal}]{barnes-etal-2019-sentiment}
\bibinfo{author}{J.~Barnes}, \bibinfo{author}{L.~{\O}vrelid},
  \bibinfo{author}{E.~Velldal},
\newblock \bibinfo{title}{Sentiment analysis is not solved! assessing and
  probing sentiment classification},
\newblock in: \bibinfo{booktitle}{Proceedings of the 2019 ACL Workshop
  BlackboxNLP: Analyzing and Interpreting Neural Networks for NLP},
  \bibinfo{publisher}{Association for Computational Linguistics},
  \bibinfo{address}{Florence, Italy}, \bibinfo{year}{2019}, pp.
  \bibinfo{pages}{12--23}.
%Type = Inproceedings
\bibitem[{Blitzer et~al.(2007)Blitzer, Dredze, and
  Pereira}]{blitzer-etal-2007-biographies}
\bibinfo{author}{J.~Blitzer}, \bibinfo{author}{M.~Dredze},
  \bibinfo{author}{F.~Pereira},
\newblock \bibinfo{title}{Biographies, {B}ollywood, boom-boxes and blenders:
  Domain adaptation for sentiment classification},
\newblock in: \bibinfo{booktitle}{Proceedings of the 45th Annual Meeting of the
  Association of Computational Linguistics}, \bibinfo{publisher}{Association
  for Computational Linguistics}, \bibinfo{address}{Prague, Czech Republic},
  \bibinfo{year}{2007}, pp. \bibinfo{pages}{440--447}.
%Type = Inproceedings
\bibitem[{Barzamini et~al.(2022)Barzamini, Rahimi, Shahzad, and
  Alhoori}]{improvingdomainspecs}
\bibinfo{author}{H.~Barzamini}, \bibinfo{author}{M.~Rahimi},
  \bibinfo{author}{M.~Shahzad}, \bibinfo{author}{H.~Alhoori},
\newblock \bibinfo{title}{Improving generalizability of ml-enabled software
  through domain specification},
\newblock in: \bibinfo{booktitle}{Proceedings of the 1st International
  Conference on AI Engineering: Software Engineering for AI}, CAIN '22,
  \bibinfo{publisher}{Association for Computing Machinery},
  \bibinfo{address}{New York, NY, USA}, \bibinfo{year}{2022}, p.
  \bibinfo{pages}{181–192}.
%Type = Article
\bibitem[{Chen et~al.(2018)Chen, Kuo, Liu, Poon, Towey, Tse, and
  Zhou}]{MTsurvey18}
\bibinfo{author}{T.~Y. Chen}, \bibinfo{author}{F.-C. Kuo},
  \bibinfo{author}{H.~Liu}, \bibinfo{author}{P.-L. Poon},
  \bibinfo{author}{D.~Towey}, \bibinfo{author}{T.~H. Tse},
  \bibinfo{author}{Z.~Q. Zhou},
\newblock \bibinfo{title}{Metamorphic testing: A review of challenges and
  opportunities},
\newblock \bibinfo{journal}{ACM Comput. Surv.} \bibinfo{volume}{51}
  (\bibinfo{year}{2018}).

\end{thebibliography}
